\pdfoutput=1

\documentclass[11pt]{article}

\usepackage[preprint]{acl}

\usepackage{times}
\usepackage{latexsym}

\usepackage[T1]{fontenc}

\usepackage[utf8]{inputenc}

\usepackage{microtype}

\usepackage{inconsolata}

\usepackage{graphicx}
\usepackage{booktabs}
\usepackage{amsmath}

\usepackage[table]{xcolor}

\usepackage[dvipsnames]{xcolor}
\usepackage{fontawesome5}

\definecolor{llmclirGrey}{RGB}{217, 217, 217}
\definecolor{llmclirBlue}{RGB}{66, 133, 244}

\usepackage{arydshln}

\title{Evaluating Large Language Models for Cross-Lingual Retrieval} 

\author{
\textbf{Longfei Zuo\thanks{\; Equal contribution.}\textsuperscript{\faMountain}} \quad
 \textbf{Pingjun Hong\footnotemark[1]\textsuperscript{\faMountain}} \quad
 \textbf{Oliver Kraus\textsuperscript{\faMountain\kern1pt}} \quad
 \\
 \textbf{Barbara Plank\textsuperscript{\faMountain\kern1pt\faRobot}} \quad
 \textbf{Robert Litschko\textsuperscript{\faMountain\kern1pt\faRobot}}
\\
\textsuperscript{\faMountain} MaiNLP, Center for Information and Language Processing, LMU Munich, Germany \\
\textsuperscript{\faRobot} Munich Center for Machine Learning (MCML), Munich, Germany \\
\tt{
\{\href{mailto:zuo.longfei@campus.lmu.de}{\textcolor{black}{zuo.longfei}},
\href{mailto:pingjun.hong@campus.lmu.de}{\textcolor{black}{pingjun.hong}}\}@campus.lmu.de},\\
\tt{
\{\href{mailto:o.kraus2n@lmu.de}{\textcolor{black}{o.kraus2}},
\href{mailto:b.plank@lmu.de}{\textcolor{black}{b.plank}}, 
\href{mailto:robert.litschko@lmu.de}{\textcolor{black}{robert.litschko}}\}@lmu.de}
}

\begin{document}

\maketitle

\begin{abstract}
Multi-stage information retrieval (IR) has become a widely-adopted paradigm in search. While Large Language Models (LLMs) have been extensively evaluated as second-stage reranking models for monolingual IR, a systematic large-scale comparison is still lacking for cross-lingual IR (CLIR). Moreover, while prior work shows that LLM-based rerankers improve CLIR performance, their evaluation setup relies on lexical retrieval with machine translation (MT) for the first stage. This is not only prohibitively expensive but also prone to error propagation across stages. Our evaluation on passage-level and document-level CLIR reveals that further gains can be achieved with multilingual bi-encoders as first-stage retrievers and that the benefits of translation diminishes with stronger reranking models. We further show that pairwise rerankers based on instruction-tuned LLMs perform competitively with listwise rerankers. To the best of our knowledge, we are the first to study the interaction between retrievers and rerankers in two-stage CLIR with LLMs. Our findings reveal that, without MT, current state-of-the-art rerankers fall severely short when directly applied in CLIR. 
\end{abstract}

\section{Introduction}

Cross-lingual information retrieval (CLIR) aims to retrieve documents written in a different language than the query, which facilitates multilingual access to information. Traditionally, CLIR systems have relied heavily on machine translation (MT) to convert either queries or documents into a shared language, effectively transforming the original cross-lingual task into a monolingual task \cite{oard-1998-comparative, mccarley-1999-translate, zhou2012, sun-etal-2020-clireval, lawrie2022hc4}. While this MT-based setting has been the dominant approach, it introduces several practical and methodological limitations: translation increases the query latency; it remains unavailable or unreliable for many low-resource languages \citep{10.1162/coli_a_00446}, adversely affecting cross-lingual retrieval when translations contain errors \citep{litschko2022parameter,guo2024query}.

\begin{figure}[t]
    \centering 
    \includegraphics[width=\columnwidth]{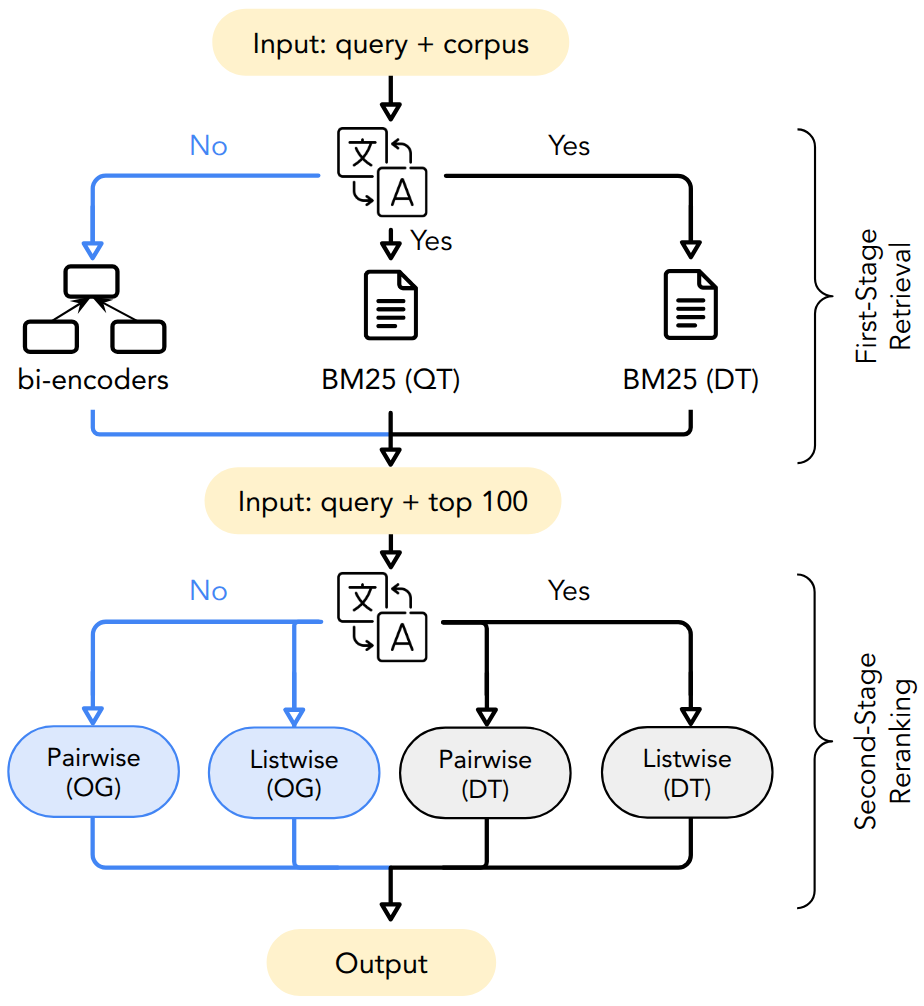}
    \caption{The pipeline used in our experiments. QT: monolingual setup using query translation; DT: monolingual setup using document translation; OG: cross-lingual setup using original documents. Routes marked with \textcolor{llmclirBlue}{blue} indicate the cross-lingual information retrieval, where translation is omitted in both the retrieval and reranking stages. Abbreviations used consistently across tables and figures.}
    \label{fig:pipeline}
\end{figure}

Recent work leveraging large language models (LLMs) in information retrieval has demonstrated promising gains over baseline systems \cite{repllama, ma2023zeroshotlistwisedocumentreranking}, highlighting their capability in ranking tasks. While prior work focuses on monolingual retrieval and reranking, cross-lingual LLM-based retrieval has been understudied. To the best of our knowledge, the only prior works on LLM-based CLIR rely on MT to bridge the language gap in the retrieval stage \citep{adeyemi-etal-2024-zero} and evaluates rerankers on a limited number of languages \citep{weller2025mfollowir}. In our study, we move beyond translation-based setups and conduct a large-scale evaluation of LLMs for cross-lingual retrieval and reranking. Figure \ref{fig:pipeline} illustrates our experimental setup. We compare the performance of state-of-the-art bi-encoders against MT-based lexical retrieval, and quantify the translation gap between rerankers applied on original language documents (OG) versus translated documents (DT). Our work is also the first to compare listwise and pairwise LLM-based rerankers on CLIR.

Our research addresses the following key questions: \textbf{RQ1:} How do recent dense multilingual bi-encoders and sparse BM25 differ in their performance as first-stage retrievers for cross-lingual retrieval?
\textbf{RQ2:} To what extent do LLM rerankers improve retrieval performance when paired with different types of first-stage retrievers, and how does this interaction vary across high-resource and low-resource language settings? 
\textbf{RQ3:} How do pairwise and listwise reranking approaches influence cross-lingual reranking performance?
\textbf{RQ4:} What is the impact of document length on listwise and pairwise reranking in CLIR?

By addressing these questions, we provide a comprehensive view of LLM rerankers’ cross-lingual capabilities and limitations, offering critical guidance for building more effective and adaptable multilingual retrieval solutions. In summary, our main contribution is a systematic evaluation of LLMs for cross-lingual reranking without fully relying on document translation. We release our code and resources to facilitate reproducibility and future research.\footnote{\url{https://github.com/mainlp/llm-clir}} 

\section{Related Work}

\subsection{Multi-Stage Retrieval}

The multi-stage retrieval paradigm, widely used in prior work \cite{nogueira2019multistagedocumentrankingbert, repllama, Zhuang_2024, rathee2025guidingretrievalusingllmbased}, consists of a fast first-stage retriever followed by one or more reranking stages for improved precision \cite{nogueira2019multistagedocumentrankingbert}. 

Prior studies have shown that first-stage quality can significantly impact reranking performance \cite{pradeep2023rankvicunazeroshotlistwisedocument, pradeep2023rankzephyreffectiverobustzeroshot}, and that reranking a smaller, high-quality candidate set (e.g., top-20) can match or exceed reranking larger document pools (e.g., top-100). Lexical retrievers like BM25 \cite{bm25} with document translation pipelines have been widely used in the first stage. More recent approaches employ bi-encoder models to improve candidate quality \cite{Liu2024LeveragingPE, Lawrie2024PLAIDSF}. Our work presents the first large-scale comparative study between LLM-based retrievers and rerankers, across different retrieval paradigms and language pairs.

Recently, many studies have leveraged LLMs as second-stage rerankers due to their strong zero-shot ranking capabilities \cite{sun-etal-2023-chatgpt, pradeep2023rankvicunazeroshotlistwisedocument, pradeep2023rankzephyreffectiverobustzeroshot, zhang2023rankwithoutgptbuildinggptindependentlistwise}. Despite their effectiveness in ranking, applying computationally expensive LLM-based rerankers over the full candidate pool is slow and resource-intensive in practice. This makes multi-stage IR crucial to first filter a manageable subset of candidates for reranking.

\subsection{Reranking Approaches with LLMs}

\paragraph{Listwise Reranking} allows LLMs to evaluate multiple candidate documents at once within a single prompt \cite{ma2023zeroshotlistwisedocumentreranking}. Given a query \(q\) and a candidate set of documents \(D_{1}, D_{2},...,D_{n}\), the LLM returns a permutation (i.e., ranked list) based on estimated relevance $D_{\pi(1)}, D_{\pi(2)}, ..., D_{\pi(n)}$. 

RankGPT, introduced by \citet{sun-etal-2023-chatgpt}, leverages instruction-tuned LLMs for passage reranking and employs a novel prompting-based strategy to enhance listwise reranking performance. However, a key limitation of this work is its reliance on proprietary models. 
Recent open-source listwise reranking models include RankVicuna \citep{pradeep2023rankvicunazeroshotlistwisedocument}, RankZephyr \citep{pradeep2023rankzephyreffectiverobustzeroshot} and Rank-without-GPT \citep{zhang2023rankwithoutgptbuildinggptindependentlistwise} have been distilled from closed-source models and shown to perform competitive performance on benchmarks like TREC DL \citep{craswell2020overviewtrec2019deep,craswell2021overviewtrec2020deep} and BEIR \citep{thakur2021beir}. However, the performance of LLM-based rerankers on cross-lingual reranking tasks is not well understood, which is a gap that our work aims to address.

\citet{adeyemi-etal-2024-zero} evaluates RankGPT and RankZephyr on low-resource English-to-African language pairs. However, their evaluation is limited to listwise reranking models with translation-based first-stage retrievers. In this study, we compare their performance against pairwise rerankers and the interaction between rerankers and lexical and dense first-stage retrieval models. Our study involves cross-lingual language pairs on both high- and low-resources languages, as well as language pairs that do not involve English queries.

\paragraph{Pairwise Reranking} frames the reranking task as presenting a query along with two candidate documents to a LLM, and the LLM is then prompted to compare the relevance between the candidates in order to select the more relevant one \cite{qin-etal-2024-large}. In their work, the authors evaluate the proposed pairwise reranking approach on variants of the encoder-decoder FLAN-T5 models \citep{tay2023ul,chung2024scaling} on monolingual retrieval benchmarks. We extend this to the cross-lingual domain and evaluate pairwise reranking on two recent multilingual decoder-only LLMs. 

\paragraph{Pointwise Reranking} is another way to rerank the candidates via relevance generation \cite{nogueira-etal-2020-document, liang2023holisticevaluationlanguagemodels, repllama} or query generation strategy \cite{sachan-etal-2022-improving}. \citet{weller2025mfollowir} provide a cross-lingual reranking evaluation of Mistral-7B-Instruct models \citep{jiang2023mistral7b,weller-etal-2025-followir} as pointwise rerankers, following the MonoT5 approach \citep{nogueira-etal-2020-document}. Since research has demonstrated that pairwise methods generally outperform pointwise approaches \cite{qin-etal-2024-large, liu2025listconrankercontrastivetextreranker}, we exclusively adopt the pairwise and listwise strategies in our experiments.

\section{Evaluation Framework}

In the following, we describe our evaluation framework shown in Figure~\ref{fig:pipeline}. To quantify to what extent gains in the retrieval stage translate to better reranking results, we experiment with translation-based lexical retrieval as well as multilingual bi-encoders. We also compare listwise and pairwise rerankers and quantify the impact of translation. 

\subsection{Datasets}

We select the 2003 portion of the \textbf{CLEF} benchmark, due to its well-established cross-lingual test collections comprising both query sets and document corpora \cite{clef2003}. In CLEF 2003 documents consist of newswire articles and cover European languages. Following the experimental setup of \citet{litschko2021crosslingualretrievalmultilingualtext}, we evaluate cross-lingual retrieval experiments across 9 language pairs: EN-\{FI, DE, IT, RU\}, DE-\{FI, IT, RU\}, and FI-\{IT, RU\}, with 60 parallel queries. 

Following \citet{adeyemi-etal-2024-zero}, we use the ``Test Set A'' split of the \textbf{CIRAL} benchmark \cite{ciral} to evaluate CLIR on low-resource African languages. CIRAL is a cross-lingual passage retrieval benchmark where queries are written in English and documents written in Hausa (HA), Somali (SO), Swahili (SW) and Yoruba (YO). Dataset statistics are shown in Table~\ref{tab:dataset-stats}. CIRAL documents are extracted from African news and blog website and chunked into passages. The average length of CIRAL passages is with 139 white-space-delimited tokens less than that of CLEF documents, which consists of 274 tokens on average. In Section~\ref{ssec:doc_length} we investigate the impact of document lengths on LLM-based reranking.

\begin{table}[t]
\centering
\footnotesize 
\setlength{\tabcolsep}{2.8pt}
\begin{tabular}{@{}lccclccc@{}}
\toprule
\multicolumn{4}{c}{\textbf{CLEF 2003}} & \multicolumn{4}{c}{\textbf{CIRAL (Test Set A)}} \\
\cmidrule(lr){1-4} \cmidrule(lr){5-8}
\textbf{Lang.} & \textbf{\#Q} & \textbf{\#D} & \textbf{Avg. Len.} & \textbf{Lang.} & \textbf{\#Q} & \textbf{\#D} & \textbf{Avg. Len.} \\

\midrule

DE & 60 & 295k & 284 & Ha & 80 & 715k & 135 \\
FI & 60 & 55k & 256 & So & 99 & 827k & 126 \\
IT & 60 & 158k & 298 & Sw & 85 & 949k & 127 \\
RU & 37 & 17k &258 & Yo & 100 & 82k & 168 \\
EN & 60 & 169k & 509 & -- & -- & -- & -- \\
\bottomrule
\end{tabular}
\caption{Statistics of the CLEF 2003 and CIRAL (Test Set A) datasets. \#Q: number of queries per language; \#D: number of documents per language; Avg. Len.: average number of  document tokens using a whitespace tokenizer.}
\label{tab:dataset-stats}
\end{table}

\subsection{Multi-Stage Pipeline: Experimental Setup}

Building upon the limitations identified in prior translation-dependent setups, we design a multi-stage retrieval pipeline that evaluates multilingual capabilities more directly. All the evaluation experiments are run on NVIDIA A100 GPUs. A summary of all models used in this study can be found in Table~\ref{tab:model_stats}.

\paragraph{First-Stage Retrieval.} As a lexical baseline retrieval method, we use BM25 via Pyserini \cite{pyserini}, with batch size of 1 and a single thread. BM25 parameters are set to $k_1 = 0.9$, $b = 0.4$. We index each collection by language and retrieve the top 100 documents for each query. Since lexical retrieval is not well-suited for cross-lingual retrieval, we index documents after translating them to the query language (see Section~\ref{ssec:doc_translation}). 

Furthermore, we chose to evaluate five state-of-the-art bi-encoder models as first-stage retrievers, divided into encoder-only and decoder-only transformer architectures. For each model we retrieve the top 100 documents with a maximum document length of 512, which is a common default value in most models. The evaluated encoder-only models include M3 \cite{chen-etal-2024-m3}, mGTE \cite{zhang-etal-2024-mgte} and multilingual E5 \cite{wang2024multilinguale5textembeddings}.
M3 is initialized from XLM-R \citep{conneau-etal-2020-unsupervised} and supports dense, sparse and multi-vector representations. In our multi-stage pipeline, we utilize the dense representations. mGTE builds on a modernized BERT which is trained from scratch on multilingual datasets. This resulting model is then further optimized for retrieval tasks on a contrastive loss objective. E5 is initialized from XLM-R and then finetuned on a contrastive loss objective. Among the available small, base, and large variants, we use multilingual-e5-large.

The evaluated decoder-only models include RepLLaMA \cite{repllama} and NV-Embed-v2 \cite{lee2025nvembed}. RepLLaMA is initialized from LLaMA-2-7B \citep{touvron2023llama} and then further finetuned for retrieval tasks. As the model is restricted to unidirectional attention, the representation of a document or query is extracted from an appended end-of-sequence token. NV-Embed-v2 is initialized from Mistral-7B \citep{jiang2023mistral7b}.  During training with a contrastive loss objective, the causal attention mask is removed, enabling bi-directional attention. Embeddings are extracted using a latent attention pooling mechanism rather than relying on a single token embedding. A practical consideration when using decoder-only models is their relatively high storage requirements. Both RepLLaMA and NV-Embed-v2 utilize a 4096-dimensional embedding space, which can lead to significantly larger index sizes and increased memory requirements compared to typical 768- or 1024-dimensional embeddings used in encoder-only models.

Our model selection was informed by their strong performance on existing monolingual benchmarks \citep{thakur2021beir,craswell2020overviewtrec2019deep,craswell2021overviewtrec2020deep} and the multilingual MMTEB benchmark \citep{enevoldsenmmteb}. By evaluating those models on the CLEF and CIRAL datasets, we shed light on their performance on typologically diverse language pairs including low-resource languages. 

\setlength{\tabcolsep}{4.9pt}
\begin{table*}[!ht]
\centering
\small 
\begin{tabular}{lllllllllll}
\toprule
\textbf{} & EN-FI & EN-IT & EN-RU & EN-DE & DE-FI & DE-IT & DE-RU & FI-IT & FI-RU & AVG \\
\hline 
\rowcolor{llmclirGrey} \multicolumn{11}{l}{\textit{First-stage retrieval (dense)}\rule{0pt}{8pt}} \\
\\[-8pt]
(1a) mGTE & \textbf{0.324} & 0.302 & 0.263 & 0.348 & \textbf{0.330} & 0.307 & \textbf{0.295} & 0.262 & 0.229 & 0.296 \\
(1b) RepLLaMA & 0.298 & 0.320 & 0.305 & 0.288 & 0.271 & 0.286 & 0.265 & 0.273 & 0.160 & 0.274 \\
(1c) M3 & 0.309 & 0.321 & 0.269 & 0.298 & 0.290 & 0.287 & 0.225 & 0.277 & 0.169 & 0.272 \\
(1d) E5 & 0.166 & 0.223 & 0.048 & 0.193 & 0.211 & 0.261 & 0.114 & 0.279 & 0.131 & 0.181 \\
(1e) NV-Embed-v2 & 0.286 & \textbf{0.450} & \textbf{0.324} & \textbf{0.422} & 0.148 & \textbf{0.404} & 0.287 & \textbf{0.342} & \textbf{0.244} & \textbf{0.323} \\
\textcolor{gray}{(1f)} \textcolor{gray}{NV-Embed-v2 \textit{(oracle)}} & 
\textcolor{gray}{0.499} & \textcolor{gray}{0.738} & \textcolor{gray}{0.545} &
\textcolor{gray}{0.683} & \textcolor{gray}{0.485} & \textcolor{gray}{0.684} &
\textcolor{gray}{0.539} & \textcolor{gray}{0.624} & \textcolor{gray}{0.477} & \textcolor{gray}{0.586} \\
\\[-8pt]

\rowcolor{llmclirGrey} \multicolumn{11}{l}{\textit{First-stage retrieval (sparse)}\rule{0pt}{8pt}} \\
\\[-8pt]
(2a) BM25-QT & 0.309 & \textbf{0.409} & 0.237 & 0.207 & 0.267 & \textbf{0.397} & \textbf{0.241} & \textbf{0.341} & \textbf{0.261} & 0.297 \\
(2b) BM25-DT & \textbf{0.413} & 0.396 & \textbf{0.255} & \textbf{0.485} & \textbf{0.301} & 0.282 & 0.216 & 0.245 & 0.179 & \textbf{0.308} \\
\textcolor{gray}{(2c)} \textcolor{gray}{BM25-DT \textit{(oracle)}} & 
\textcolor{gray}{0.613} & \textcolor{gray}{0.686} & \textcolor{gray}{0.571} &
\textcolor{gray}{0.716} & \textcolor{gray}{0.617} & \textcolor{gray}{0.533} &
\textcolor{gray}{0.481} & \textcolor{gray}{0.446} & \textcolor{gray}{0.378} & \textcolor{gray}{0.560} \\
\\[-8pt]

\hline

\rowcolor{llmclirGrey} \multicolumn{11}{l}{\textit{Listwise Reranking (Retriever: NV-Embed-v2)}\rule{0pt}{8pt}} \\
\\[-8pt]
(3a) RankZephyr (OG) & 0.351$^{*}$ & 0.453 & 0.340 & 0.444 & 0.287$^{*}$ & 0.385 & 0.309 & 0.300 & 0.250 & 0.347 \\
(3b) RankGPT$_{3.5}$ (OG) & 0.337 & 0.438 & 0.325 & 0.438 & 0.297$^{*}$ & 0.428 & 0.305 & 0.386$^{*}$ & 0.250 & 0.356 \\
(3c) RankGPT$_{4.1}$ (OG) & 0.416$^{*}$ & 0.526$^{*}$ & \textbf{0.417$^{*}$} & 0.506$^{*}$ & 0.400$^{*}$ & 0.501$^{*}$ & \textbf{0.410$^{*}$} & 0.449$^{*}$ & 0.339 & 0.440 \\ 
\cdashline{1-11}[.4pt/1pt]\noalign{\vskip 0.5ex}
(3d) RankZephyr (DT) & 0.402$^{*}$ & 0.466 & 0.375 & 0.472$^{*}$ & 0.347$^{*}$ & 0.394 & 0.318 & 0.308 & 0.217 & 0.367 \\
(3e) RankGPT$_{3.5}$ (DT) & 0.383$^{*}$ & 0.472 & 0.308 & 0.450 & 0.311$^{*}$ & 0.432 & 0.336 & 0.375 & 0.239 & 0.367\\
(3f) RankGPT$_{4.1}$ (DT) & \textbf{0.433}$^{*}$ & \textbf{0.557$^{*}$} & 0.382 & \textbf{0.517$^{*}$} & \textbf{0.402$^{*}$} & \textbf{0.518$^{*}$} & 0.363$^{*}$ & \textbf{0.472$^{*}$} & \textbf{0.341} & \textbf{0.443}\\
\\[-8pt]

\rowcolor{llmclirGrey} \multicolumn{11}{l}{\textit{Listwise Reranking (Retriever: BM25-DT)}\rule{0pt}{8pt}} \\
\\[-8pt]
(4a) RankZephyr (OG) &0.402 & 0.436 & 0.304 & 0.473 & 0.349 & 0.350$^{*}$ & 0.315$^{*}$ & 0.272 & 0.176 & 0.342 \\
(4b) RankGPT$_{3.5}$ (OG) & 0.405 & 0.423 & 0.283 & 0.479 & 0.389$^{*}$ & 0.338$^{*}$ & 0.259 & 0.288$^{*}$ & 0.190 & 0.339 \\
(4c) RankGPT$_{4.1}$ (OG) & 0.480$^{*}$ & 0.512$^{*}$ & \textbf{0.402$^{*}$} & 0.544$^{*}$ & 0.465$^{*}$ & \textbf{0.421$^{*}$} & 0.318$^{*}$ & \textbf{0.355$^{*}$} & 0.244 & 0.416 \\ 
\cdashline{1-11}[.4pt/1pt]\noalign{\vskip 0.5ex} 
(4d) RankZephyr (DT) & 0.461 & 0.475$^{*}$ & 0.333$^{*}$ & 0.510 & 0.425$^{*}$ & 0.366$^{*}$ & 0.322$^{*}$ & 0.230 & 0.190 & 0.368 \\
(4e) RankGPT$_{3.5}$ (DT) & 0.405 & 0.467$^{*}$ & 0.303$^{*}$ & 0.487 & 0.393$^{*}$ & 0.361$^{*}$ & 0.278 & 0.270 & 0.184 & 0.350\\
(4f) RankGPT$_{4.1}$ (DT) & \textbf{0.505$^{*}$} & \textbf{0.537$^{*}$} & 0.373$^{*}$ & \textbf{0.554$^{*}$} & \textbf{0.477$^{*}$} & 0.413$^{*}$ & \textbf{0.341$^{*}$} & 0.341$^{*}$ & \textbf{0.260} & \textbf{0.422}\\

\bottomrule
\end{tabular}
\caption{MAP scores of first-stage retriever and listwise reranking results on CLEF 2003, with the best performance for each language pair marked in \textbf{Bold}. \textcolor{gray}{Gray font} indicates the best possible (oracle) reranking performance achievable based on the top-100 documents provided by the first-stage retriever. OG denotes cross-lingual reranking with documents in their original language. QT and DT denote experiments involving query and document translation. $^*$: statistically significant difference to the first-stage retriever (paired t-test, $p<0.05$).
}
\label{tab:clef_list}
\end{table*}

\paragraph{Listwise Reranking.} We include listwise reranking in the second stage of the retrieval pipeline due to its demonstrated efficacy \cite{sun-etal-2023-chatgpt}. Specifically, we employ RankZephyr and RankGPT$_{3.5}$ following \citet{adeyemi-etal-2024-zero}, and additionally incorporate RankGPT$_{4.1}$ into our evaluation. RankZephyr performs well in both monolingual \citep{pradeep2023rankzephyreffectiverobustzeroshot} and MT-based cross-lingual retrieval \cite{adeyemi-etal-2024-zero}, but its effectiveness in cross-lingual reranking without MT—where both query and documents are non-English—remains underexplored. We evaluate RankZephyr using the \texttt{rank\_llm} framework \citep{sharifymoghaddam2025rankllm},\footnote{\url{https://github.com/castorini/rank_llm}} with a sliding window size of 20, step size of 10, and a max context length of 4096 tokens. The reranking prompt is shown in Figure \ref{fig:listwise-prompt} in Appendix~\ref{sec:prompt}.

\paragraph{Pairwise Reranking.} In our experiments, we utilized the Pairwise Ranking Prompting (PRP) with the bubble-sort-like sliding window strategy \citep{qin-etal-2024-large}, in which \(k\) passes are performed by comparing each document pair from bottom to the top with the initial ranking. Following the prompt design of \citet{qin-etal-2024-large} (see Figure \ref{fig:pairwise-prompt}), we perform $k=10$ passes through the candidate list. We adopt the generation mode, where models output either ``Passage A'' or ``Passage B''. The experiments are conducted using two multilingual LLMs: Llama-3.1-8B-Instruct \cite{grattafiori2024llama3herdmodels}  and Aya-101 \cite{ustun-etal-2024-aya}, due to their strong multilingual capabilities. While Aya-101 has seen all four CIRAL and five CLEF languages, Llama-3.1-8B-Instruct only supports three of those languages (English, German, and Italian). 

\subsection{Document Translation Setup}
\label{ssec:doc_translation}

To compare cross-lingual and monolingual retrieval performance, we adopt a translation-based setup that converts the task into a monolingual one, isolating the impact of language mismatch. 

To maintain consistency with the official translations of the CIRAL dataset \cite{ciral}---which are generated using the nllb-200-1.3B model \citep{team2022NoLL}, we also translate the entire CLEF 2003 documents collection for all nine language pairs using the same model. In order to ensure a high translation quality, we use a sentence-level strategy: documents are split into sentences, translated individually, and then concatenated to reconstruct the full document. In a preliminary study, we found that this approach improves the translation accuracy and avoids errors that often arise in document-level translation. In particular, we found that sentence-by-sentence translations reduces word and phrase repetitions, which occurred more often when documents were translated as a whole. All sentence translations use a maximum input sequence length of 128 and batch size of 256 instances. Translations are generated using greedy decoding. 

\setlength{\tabcolsep}{12pt}
\begin{table*}[!ht]
\centering
\small 
\begin{tabular}{llllll}
\toprule
\textbf{} & EN-HA & EN-SO & EN-SW & EN-YO & AVG \\
\hline 
\rowcolor{llmclirGrey}\multicolumn{6}{l}{\textit{First-stage retrieval (dense)}\rule{0pt}{8pt}} \\
\\[-8pt]
(1a) mGTE & 0.252 & 0.266 & 0.317 & 0.339 & 0.294 \\
(1b) RepLLaMA & 0.199 & 0.192 & 0.183 & 0.315 & 0.222 \\
(1c) E5 & 0.291 & 0.278 & 0.326 & 0.415 & 0.327 \\
(1d) NV-Embed-v2 & 0.136 & 0.263 & 0.290 & \textbf{0.471} & 0.290 \\
(1e) M3 & \textbf{0.388} & \textbf{0.351} & \textbf{0.402} & 0.425 & \textbf{0.392} \\
\textcolor{gray}{(1f)} \textcolor{gray}{M3 \textit{(oracle)}} & \textcolor{gray}{0.744} & \textcolor{gray}{0.687} & \textcolor{gray}{0.792} & \textcolor{gray}{0.793} & \textcolor{gray}{0.754}  \\
\\[-8pt]

\rowcolor{llmclirGrey}\multicolumn{6}{l}{\textit{First-stage retrieval (sparse)}\rule{0pt}{8pt}} \\
\\[-8pt]
(2a) BM25-QT & 0.087 & 0.081 & 0.130 & 0.286 & 0.146 \\
(2b) BM25-DT & \textbf{0.214} & \textbf{0.246} & \textbf{0.233} & \textbf{0.445} & \textbf{0.285} \\
\textcolor{gray}{(2c)} \textcolor{gray}{BM25-DT \textit{(oracle)}} & \textcolor{gray}{0.586} & \textcolor{gray}{0.561} & \textcolor{gray}{0.611} & \textcolor{gray}{0.826} & \textcolor{gray}{0.646}  \\
\\[-8pt]

\hline 

\rowcolor{llmclirGrey}\multicolumn{6}{l}{\textit{Listwise Reranking (Retriever: M3)}\rule{0pt}{8pt}} \\
\\[-8pt]
(3a) RankZephyr (OG) & 0.352 & 0.302 & 0.372 & 0.433 & 0.365 \\
(3b) RankGPT$_{3.5}$ (OG) & 0.419$^{*}$ & 0.382$^{*}$ & 0.413 & 0.484$^{*}$ & 0.425 \\
(3c) RankGPT$_{4.1}$ (OG) & 0.467$^{*}$ & 0.453$^{*}$ & 0.485$^{*}$ & 0.566$^{*}$ & 0.493 \\ 
\cdashline{1-6}[.4pt/1pt]\noalign{\vskip 0.5ex}
(3d) RankZephyr (DT) & 0.464$^{*}$ & 0.454$^{*}$ & 0.448$^{*}$ & 0.540$^{*}$ & 0.477 \\ 
(3e) RankGPT$_{3.5}$ (DT) & 0.439$^{*}$ & 0.395$^{*}$ & 0.419 & 0.491$^{*}$ & 0.436 \\
(3f) RankGPT$_{4.1}$ (DT) & \textbf{0.490$^{*}$} & \textbf{0.481$^{*}$} & \textbf{0.498$^{*}$} & \textbf{0.576$^{*}$} & \textbf{0.511} \\
\\[-8pt]

\rowcolor{llmclirGrey}\multicolumn{6}{l}{\textit{Listwise Reranking (Retriever: BM25-DT)}\rule{0pt}{8pt}} \\
\\[-8pt]

(4a) RankZephyr (OG) & 0.260$^{*}$ & 0.300$^{*}$ & 0.291$^{*}$ & 0.439 & 0.322 \\
(4b) RankGPT$_{3.5}$ (OG) & 0.241 & 0.292 & 0.256 & 0.442 & 0.308 \\
(4c) RankGPT$_{4.1}$ (OG) & 0.383$^{*}$ & 0.354$^{*}$ & 0.361$^{*}$ & 0.574$^{*}$ & 0.418 \\ 
\cdashline{1-6}[.4pt/1pt]\noalign{\vskip 0.5ex}
(4d) RankZephyr (DT) & 0.371$^*$ & 0.362$^*$ & 0.365$^*$ & 0.531$^*$ & 0.407 \\
(4e) RankGPT$_{3.5}$ (DT) &0.298 & 0.308 & 0.307 & 0.499 & 0.353 \\
(4f) RankGPT$_{4.1}$ (DT) & \textbf{0.397$^{*}$} & \textbf{0.378$^{*}$} & \textbf{0.406$^{*}$} & \textbf{0.584$^{*}$} & \textbf{0.441} \\

\bottomrule
\end{tabular}
\caption{nDCG@20 scores of first-stage retrievers and listwise rerankers on the CIRAL dataset, with the best performance for each language pair marked in \textbf{Bold}. \textcolor{gray}{Gray font} indicates the best possible (oracle) reranking results achievable based on the top-100 documents provided by the first-stage retriever. $^*$: statistically significant difference to the retriever (paired t-test, $p<0.05$). Rows 2a and 4b,e are taken from \protect\citep{adeyemi-etal-2024-zero}.}
\label{tab:ciral_list}
\end{table*}

\section{Results and Discussion}
\label{sec:results}

\subsection{First-Stage Retrieval}

Table \ref{tab:clef_list} (rows 1a-1e) presents the results for the first-stage retrievers on the CLEF 2003 dataset. NV-Embed-v2 shows the best average performance (0.323 MAP) with a notable drop in performance for the DE-FI language pair (0.148 MAP), despite having been pre-trained on both languages. mGTE exhibits the second best performance, and is able to outperform NV-Embed-v2 on the two language pairs DE-FI and DE-RU. This is particularly noteworthy, as mGTE is the smallest model in terms of parameter size (see Table~\ref{tab:model_stats}). We find that E5 shows the weakest overall performance (0.181 MAP). Translation-based lexical retrieval (2a-b) outperforms all bi-encoders except for NV-Embed-v2. 

Table \ref{tab:ciral_list} (rows 1a-1e) presents the results of the first-stage retrievers on the CIRAL dataset. Here, M3 is the best-performing model, followed by E5, which is notable given its relatively poor performance on CLEF 2003. Both NV-Embed-v2 and mGTE perform substantially worse on CIRAL. We find that language coverage explains this difference only to a certain extent. While NV-Embed-v2 does not support any of the four CIRAL languages, mGTE does support Hausa. Different from our results on CLEF, we find document translation to perform substantially worse. BM25-DT falls behind most bi-encoders, with a notable gap to M3.

These results show that the best-performing first-stage ranker differs considerably based on the chosen dataset (\textbf{RQ1}). Meanwhile, it seems likely that main factors for the performance differences are the quality and amount of training data a model has seen. While architectural differences between models (within the classes of encoder-only models and decoder-only models respectively) may play a role, training data composition and quality appears to be a more significant factor. For instance, E5 and M3 share similar architectural foundations, yet their performance diverges substantially on CIRAL.

Compared to prior work that utilized cross-lingual word embeddings \citep{glavas-etal-2019-properly, litschko2021crosslingualretrievalmultilingualtext, zhou2022neural} and multilingual pre-trained language models \citep{litschko2019evaluating}, we observe that recent bi-encoders achieve superior performance.

\subsection{Second-Stage Reranking}

\paragraph{Listwise Reranking Results.} We show the results for listwise rerankers applied on CLEF and CIRAL in Tables~\ref{tab:clef_list} and \ref{tab:ciral_list}. On average, all rerankers manage to improve the results of the input rankings generated by their respective first-stage retrievers. At the level of individual language pairs, we find that only RankGPT$_{4.1}$ improves its input rankings across the board on both datasets. On the CLEF dataset, RankZephyr and RankGPT 3.5 show improved performance over the input rankings generated by NV-Embed-v2 for most language pairs (7-8 out of 9, rows 3a-b), but the improvement is less pronounced when using BM25-DT as the first-stage retriever, with 6-7 language pairs showing improvement (rows 4a-b).

Consistent across both datasets, we find that improved retrieval results translate to better reranking results (\textbf{RQ2}). Specifically, as shown in Table~\ref{tab:clef_list}, we observe that the modest improvement in retrieval results achieved by NV-Embedd-v2 over BM25-DT on the CLEF dataset (+0.015 MAP) translates to relatively small differences in reranking results, ranging from +0.024 for RankGPT$_{3.5}$ (comparing rows 3b and 4b) to +0.05 for RankZephyr (comparing rows 3a and 4a). In contrast, our results on the CIRAL dataset (Table~\ref{tab:ciral_list}) reveal a more substantial improvement in retrieval results, with M3 outperforming BM25-DT by +0.110 MAP. Consequently, the reranking improvements are more pronounced, spanning from +0.043 for RankZephyr (comparing rows 3a and 4a) to +0.117 for RankGPT$_{3.5}$. 

The differences between CLEF and CIRAL are likely the result of differences in translation qualities \citep{adeyemi-etal-2024-zero} (see also Appendix~\ref{s:translation}). Interestingly, as models achieve an overall stronger performance, the benefits from  document translation appears to diminish. On CLEF with NV-Embed-v2 as a first-stage retriever, we observe that improvements for RankZephyr and RankGPT$_{3.5}$ are with +0.026 and +0.011 MAP (comparing rows 3a-b with 3d-e) larger than for RankGPT$_{4.1}$, where it is only +0.003 (comparing 3c with 3f). We observe a similar trend on the CIRAL dataset, where the improvements resulting from document translation for RankZephyr is with +0.112 (3a, 3d) is higher than +0.018 for RankGPT$_{4.1}$ (3c, 3f).

\setlength{\tabcolsep}{4.7pt}
\begin{table*}[!ht]
\centering
\small 
\begin{tabular}{llllllllllll}
\toprule

\textbf{} & EN-FI & EN-IT & EN-RU & EN-DE & DE-FI & DE-IT & DE-RU & FI-IT & FI-RU & AVG \\
\hline 
\rowcolor{llmclirGrey} \multicolumn{11}{l}{\textit{First-stage retrieval}\rule{0pt}{8pt}} \\
\\[-8pt]

NV-Embed-v2 & 0.286 & \textbf{0.450} & 0.324 & 0.422 & 0.148 & \textbf{0.404} & \textbf{0.287} & \textbf{0.342} & \textbf{0.244} & \textbf{0.323} \\
BM25-DT & \textbf{0.413} & 0.396 & \textbf{0.255} & \textbf{0.485} & \textbf{0.301} & 0.282 & 0.216 & 0.245 & 0.179 & 0.308 \\

\hline 

\rowcolor{llmclirGrey} \multicolumn{11}{l}{\textit{Pairwise Reranking (Retriever: NV-Embed-v2)}\rule{0pt}{8pt}} \\
\\[-8pt]
Llama-3.1-8B-Instruct (OG) & 0.354$^{*}$ & 0.474 & \textbf{0.361} & 0.438$^{*}$ & 0.282$^{*}$ & 0.439$^{*}$ & 0.338$^{*}$ & \textbf{0.395}$^{*}$ & 0.254$^{*}$ & 0.371 \\
 Aya-101 (OG)  & 0.350$^{*}$ & 0.476 & 0.335 & 0.430 & 0.289$^{*}$ & 0.436 & \textbf{0.339} & 0.379$^{*}$ & \textbf{0.278} & 0.368 \\
 \cdashline{1-11}[.4pt/1pt]\noalign{\vskip 0.5ex}
 Llama-3.1-8B-Instruct (DT)  & \textbf{0.369$^{*}$} & \textbf{0.502$^{*}$} & 0.358 & \textbf{0.461$^{*}$} & \textbf{0.333$^{*}$} & \textbf{0.459$^{*}$} & 0.335 & 0.369 & 0.276 & \textbf{0.385} \\
 Aya-101 (DT)  & 0.337$^{*}$ & 0.464 & 0.326 & 0.433 & 0.301$^{*}$ & 0.434 & 0.331 & 0.352 & 0.239 & 0.357 \\
 \\[-8pt]

\rowcolor{llmclirGrey} \multicolumn{11}{l}{\textit{Pairwise Reranking (Retriever: BM25-DT)}\rule{0pt}{8pt}} \\
\\[-8pt]
Llama-3.1-8B-Instruct (OG) & 0.460$^{*}$ & 0.450$^{*}$ & \textbf{0.312$^{*}$} & 0.492 & 0.383$^{*}$ & \textbf{0.359$^{*}$} & 0.269 & \textbf{0.286$^{*}$} & 0.191 & 0.356 \\
 Aya-101 (OG) & 0.449$^{*}$ & 0.418 & 0.301 & 0.498$^{*}$ & 0.355$^{*}$ & 0.328$^{*}$ & \textbf{0.313$^{*}$} & 0.274$^{*}$ & \textbf{0.234} & 0.352 \\
 \cdashline{1-11}[.4pt/1pt]\noalign{\vskip 0.5ex}
 Llama-3.1-8B-Instruct (DT)  & \textbf{0.479$^{*}$} & \textbf{0.472$^{*}$} & 0.293 & \textbf{0.509} & 0.390$^{*}$ & 0.352$^{*}$ & 0.272 & 0.281$^{*}$ & 0.196 & \textbf{0.360} \\
 Aya-101 (DT) & 0.449 & 0.432$^{*}$ & 0.296 & 0.497 & \textbf{0.394$^{*}$} & 0.336$^{*}$ & 0.284$^{*}$ & 0.260 & 0.199 & 0.350 \\

\bottomrule
\end{tabular}
\caption{MAP scores of pairwise reranking on CLEF 2003, with the best performance for each language pair marked in \textbf{Bold}. 
$^*$: statistically significant difference to the first-stage retriever (paired t-test, $p<0.05$).}
\label{tab:clef_pair}
\end{table*}
\setlength{\tabcolsep}{12pt}
\begin{table*}[!ht]
\centering
\small 
\begin{tabular}{llllll}
\toprule
\textbf{} & EN-HA & EN-SO & EN-SW & EN-YO & AVG \\
\hline 

\rowcolor{llmclirGrey} \multicolumn{6}{l}{\textit{First-stage retrieval}\rule{0pt}{8pt}} \\
\\[-8pt]
M3 & \textbf{0.388} & \textbf{0.351} & \textbf{0.402} & 0.425 & \textbf{0.392} \\
BM25-DT & 0.214 & 0.236 & 0.233 & \textbf{0.445} & 0.282 \\

\hline 

\rowcolor{llmclirGrey} \multicolumn{6}{l}{\textit{Pairwise Reranking (Retriever: M3)}\rule{0pt}{8pt}} \\
\\[-8pt]
Llama-3.1-8B-Instruct (OG) & 0.399 & 0.360 & 0.423$^{*}$ & 0.453$^{*}$ & 0.409 \\
Aya-101 (OG)&0.427$^{*}$ & 0.387$^{*}$ & \textbf{0.437$^{*}$} & 0.483$^{*}$ & 0.434 \\
\cdashline{1-6}[.4pt/1pt]\noalign{\vskip 0.5ex}
Llama-3.1-8B-Instruct (DT) & 0.410$^{*}$ & \textbf{0.398$^{*}$} & 0.431$^{*}$ & 0.474$^{*}$ & 0.428 \\
Aya-101 (DT)& \textbf{0.431$^{*}$} & 0.395$^{*}$ & 0.432$^{*}$ & \textbf{0.505$^{*}$} & \textbf{0.441} \\
\\[-8pt]

\rowcolor{llmclirGrey} \multicolumn{6}{l}{\textit{Pairwise Reranking (Retriever: BM25-DT)}\rule{0pt}{8pt}} \\
\\[-8pt]
Llama-3.1-8B-Instruct (OG) & 0.241$^{*}$ & 0.262$^{*}$ & 0.280$^{*}$ & 0.473$^{*}$ & 0.314 \\
Aya-101 (OG) & \textbf{0.325$^{*}$} & \textbf{0.317$^{*}$} & 0.298$^{*}$ & \textbf{0.504$^{*}$} & \textbf{0.361} \\
 \cdashline{1-6}[.4pt/1pt]\noalign{\vskip 0.5ex}
Llama-3.1-8B-Instruct (DT) &  0.281$^{*}$ & 0.284$^{*}$ & \textbf{0.306$^{*}$} & 0.503$^{*}$ & 0.344 \\
Aya-101 (DT) & 0.303$^{*}$ & 0.314$^{*}$ & 0.305$^{*}$ & 0.501$^{*}$ & 0.356 \\
\bottomrule
\end{tabular}
\caption{nDCG@20 scores of pairwise reranking on CIRAL, with the best result for each language pair marked in \textbf{Bold}. $^*$: statistically significant difference to the first-stage retriever (paired t-test, $p<0.05$).}
\label{tab:ciral_pair}
\end{table*}

\paragraph{Pairwise Reranking Results.} Tables~\ref{tab:clef_pair} and \ref{tab:ciral_pair} show the pairwise reranking results on CLEF and CIRAL. On average across all language pairs, we find that all pairwise reranking models improve their input rankings (\textbf{RQ3}). The results on CLEF with NV-Embed-v2 as the first-stage retriever\footnote{We focus our discussion on the reranking results based on the better-performing multilingual bi-encoders. In Appendix~\ref{sec:hybrid} we show that further gains are obtainable with hybrid retrieval.} show that Llama-3.1-8B-Instruct achieves a performance of 0.371 and 0.385 with and without document translation, outperforming both RankZephyr (0.347 and 0.367) and RankGPT$_{3.5}$ (0.356 and 0.367). The Aya-101 reranker outperforms both listwise models on translated documents. 

The results on the CIRAL dataset with M3 as the first-stage retriever show that the better-performing Aya-101 reranker only outperforms RankZephyr and RankGPT$_{3.5}$ when documents remain in their original language. GPT$_{4.1}$ outperforms both pairwise models. However, it is worth noting that pairwise rerankers are only based on instruction-tuned LLMs and have not been further post-trained on reranking data. This is a key difference to listwise rerankers used in this study, which are either closed-source or have been distilled from closed-source models.

\paragraph{The Glass Ceiling of Reranking.} While improvements after second-stage reranking are commonly reported, it remains valuable to examine how closely current methods approximate the optimal ranking. To estimate the possible best result of reranking, we place all relevant documents in the retrieved candidate list at the top positions to simulate the oracle first-stage result (row 1f \& 2c in Table \ref{tab:clef_list} \& Table \ref{tab:ciral_list}). The difference between this oracle ranking result and the actual first-stage scores show best possible gains for second-stage reranking.

We define \textit{Potential Reranking Improvements (PRI)} and the \textit{Realized} improvement percentage as:
\begin{align}
\text{PRI} &= s^* - s_1 \\
\text{Realized} &= \frac{s_2 - s_1}{\text{PRI}} \times 100
\end{align}
where \( s_1 \) denotes the performance of the best dense or sparse first-stage retriever, \( s^{*} \) denotes the corresponding oracle performance achievable at the first stage, and \( s_2 \) denotes the best second-stage reranking performance based on the respective \( s_1 \).

On CLEF 2003, the PRI is 0.263 for NV-Embed-v2 and 0.252 for BM25 with document translation (DT). Scores on the CIRAL dataset are higher---0.362 for the M3 retriever and 0.361 for BM25 (DT), indicating a larger room for improvement.

In terms of realized improvements, all three rerankers enhance the first-stage results under document translation (DT) settings on both datasets. RankGPT$_{4.1}$ (DT) consistently achieves the best performance, realizing 45.6\% and 45.2\% of the PRI on the CLEF dataset when reranking NV-Embed-v2 and BM25 (DT) outputs, and 32.9\% and 43.2\% on the CIRAL dataset when reranking the M3 bi-encoder and BM25 (DT) results. However, the original setting (OG) sometimes fails to outperform the first-stage results (row 3a in Table \ref{tab:ciral_list}), suggesting that reranking may introduce additional noise in certain cross-lingual settings.

Although document translation can help narrow the performance gap toward the best possible results, the highest realized improvements still fall short of expectations. Both datasets reveal a clear ``ceiling effect'' in reranking: even state-of-the-art rerankers struggle to approach the upper bound, especially in cross-lingual settings. This highlights a substantial gap between current reranking capabilities and their theoretical potential, suggesting that much of the available improvement remains untapped.

\subsection{Impact of Document Length}
\label{ssec:doc_length}

\begin{figure}[t!]
  \includegraphics[width=0.85\linewidth]{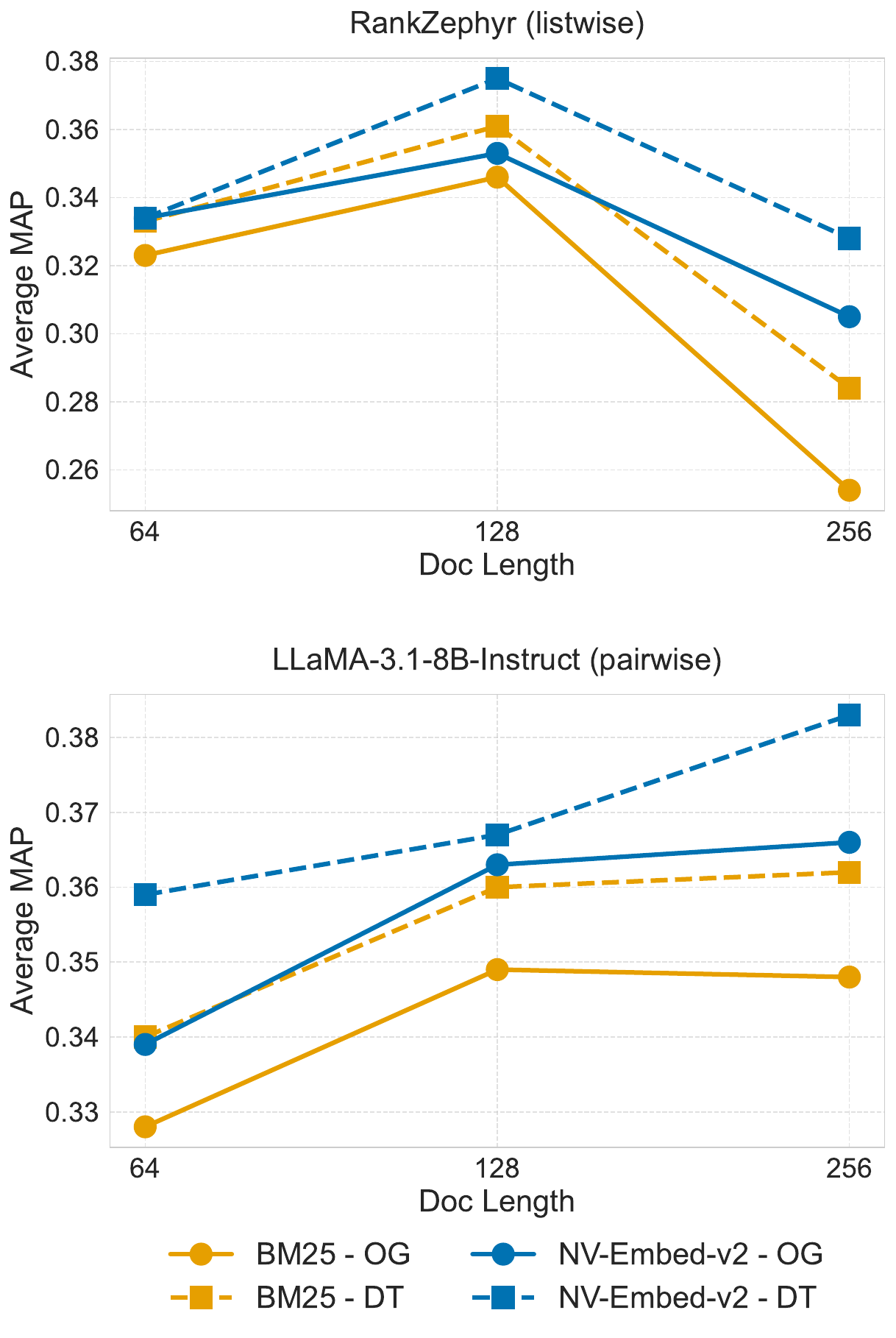}
  \caption{Effect of input document length on reranking across retrievers and reranking approaches. Tokenization is performed using each model’s own tokenizer. Results are averaged over all CLEF language pairs.}
  \label{fig:doc_length}
\end{figure}

The passage-level results on CIRAL are noticeably better than the document-level results on CLEF, which motivates us to explore the influence of documents lengths on the reranking performance.

We conduct an ablation study on the CLEF 2003 dataset by varying the maximum number of input tokens per document chunk in both listwise and pairwise setups. This parameter plays a crucial role in balancing information sufficiency and information overload for the reranker.

As shown in Figure~\ref{fig:doc_length}, RankZephyr reranker performs best at medium input lengths (128 tokens), with a noticeable performance drop at 256 tokens across both sparse (BM25) and dense (NV-Embed-v2) retrievers. This suggests that listwise rerankers are sensitive to overly long inputs, which may dilute useful signals with unnecessary context. In contrast, Llama-based pairwise rerankers show more stable or even improving trends. When using NV-Embed-v2 as the retriever, performance increases steadily from 64 to 256 tokens, indicating that this setup benefits from additional context. However, when using BM25, the gains from longer input saturate or even slightly decline at 256 tokens for DT. This suggests that the pairwise reranker, when paired with high-quality dense retriever, is more robust to longer input spans. 

We hypothesize that pairwise reranking is a simpler task and may help capture more nuanced comparison between the two candidates. In this case, longer documents provide additional context, which pairwise rerankers can utilize more effectively (\textbf{RQ4}). Overall, these results show that optimal input length is task- and model-dependent and may require careful tuning.

\section{Conclusion}

In this paper, we conducted a systematic evaluation of LLMs for CLIR, evaluating both passage- and document-level reranking without relying entirely on machine translation. Our results reveal that further cross-lingual reranking gains can be achieved by substituting lexical MT-based retrievers with better-performing multilingual bi-encoder and hybrid retrieval approaches, and that the benefits resulting from document translation diminish with stronger listwise rerankers. We further demonstrate that instruction-tuned pairwise rerankers perform competitively with listwise rerankers like RankZephyr. Finally, we highlight the sensitivity of reranking performance to document length and language resource disparities. Our findings highlight the need for more robust approaches to harness LLMs for 
cross-lingual reranking without relying on machine translation. 

\section*{Limitations}

While our study provides a comprehensive evaluation of LLMs for CLIR, several limitations remain. First, our evaluation is limited to a fixed set of LLMs, which, while representative, may not represent the full diversity of available open-source or commercial models. Second, although we compare document and query translation strategies, translation quality remains a potential confounder, as noisy translations may affect LLM reranking behavior. Finally, we restrict our experiments to zero-shot reranking without fine-tuning or domain adaptation. This may also underestimate the true potential of rerankers in downstream applications.

\section*{Acknowledgments} 

We thank the members of the MaiNLP lab for their insightful feedback on earlier drafts of this paper. We specifically appreciate the suggestions of Beiduo Chen, Florian Eichin, Siyao (Logan) Peng and Felicia Körner. The translation and document icons used in Figure~\ref{fig:pipeline} are made by Freepik from www.flaticon.com.
BP acknowledges funding by ERC Consolidator Grant DIALECT 101043235.

\paragraph{Ethical considerations.} We do not foresee any ethical concerns associated with this work. All analyses were conducted using publicly available datasets and models. No private or sensitive information was used. Additionally, we release our code, prompts, and documentations to support transparency and reproducibility.

\paragraph{Use of AI Assistants.}The authors acknowledge the use of ChatGPT exclusively for assistance with grammar, punctuation, and vocabulary refinement, as well as for support with coding-related tasks.

\bibliography{custom}

\appendix
\clearpage

\section{Prompt Template for Listwise and Pairwise Reranking}
\label{sec:prompt}
Figures~\ref{fig:listwise-prompt}~and~\ref{fig:pairwise-prompt} show the prompt templates used in our experiments. For listwise reranking, we adopt the prompt template from the original RankZephyr implementation \cite{pradeep2023rankzephyreffectiverobustzeroshot}. For the pairwise reranking approach, we use the prompt introduced in \citep{qin-etal-2024-large} and implementation provided by \citet{Zhuang_2024}.

\begin{figure}[h!]
\centering
\small
\fcolorbox{white}{gray!8}{
  \parbox{.95\linewidth}{
    \ttfamily
    \textbf{LISTWISE RERANKING PROMPT} \\[1ex]
    <|systems|> \\
    You are RankLLM, an intelligent assistant that can rank passages based on their relevancy to the query.\\[1ex]
    <|user|> \\
    I will provide you with \{num\} passages, each indicated by a numerical identifier []. Rank the passages based on their relevance to the search query: \{query\}. \\
    \\
    \texttt{[1]} \{passage 1\} \\
    \texttt{[2]} \{passage 2\} \\
    ... \\
    \texttt{[\{num\}]} \{passage \{num\}\} \\
    \\
    Search Query: \{query\}. \\
    \\
    Rank the \{num\} passages above based on their relevance to the search query. All the passages should be included and listed using identifiers, in descending order of relevance. The output format should be [] > [], e.g., [4] > [2]. Only respond with the ranking results, do not say any word or explain. \\
    <|assistant|> \\
    \\
    {\normalfont\textit{Model Generation: [9] > [4] > [20] > ... > [13]}}
  }
}
\caption{Listwise reranking prompt template used for LLM-based reranking.}
\label{fig:listwise-prompt}
\end{figure}

\begin{figure}[h!]
\centering
\small
\fcolorbox{white}{gray!8}{
  \parbox{0.95\linewidth}{
    \ttfamily
    \textbf{PAIRWISE RERANKING PROMPT} \\[1ex]
    Given a query \{query\}, which of the following two passages is more relevant to the query?\\[1ex]
    Passage A: \{document\textsubscript{1}\}\\
    Passage B: \{document\textsubscript{2}\}\\[1ex]
    Output Passage A or Passage B:
  }
}
\caption{Pairwise reranking prompt template used for LLM-based reranking.}
\label{fig:pairwise-prompt}
\end{figure}

\section{Query and Document Translation}
\label{s:translation}

We compare the impact of query translation (QT) and document translation (DT) strategies across the two benchmarks. As shown in Table \ref{tab:clef_list} and Table \ref{tab:ciral_list}, 

While both QT and DT are considered in the first stage, we only use DT to construct the candidate pool for the second stage due to its better performance.
Notably, the performance gap between QT and DT is much larger on CIRAL than on CLEF in the first stage. QT is less effective on CIRAL, as translating queries into low-resource African languages often produces short or low-quality queries that struggle to match the document content. 
By contrast, QT performs better on CLEF, where queries are translated into higher-resourced languages with more reliable translation and greater lexical overlap.

For the reranking stage, applying DT on CIRAL has a much more pronounced effect than on CLEF, resulting in a large performance gap between DT and OG. We attribute this to two factors: (1) translating documents into English transforms the reranking task into a noisy EN–EN format, which aligns more closely with the LLM's training distribution, and (2) CIRAL documents are relatively short, leading to higher translation accuracy and less noise compared to longer texts.

In contrast, the OG setting on CLEF is less problematic. The languages included are better supported by existing LLMs, and the queries themselves are typically longer and more descriptive (both title and description are included in queries), making them easier to interpret and translate.  As a result, the performance gap between OG and DT is smaller. 

\setlength{\tabcolsep}{4.7pt}
\begin{table*}[ht!]
\centering
\small 
\begin{tabular}{lllllllllll}
\toprule
\textbf{} & EN-FI & EN-IT & EN-RU & EN-DE & DE-FI & DE-IT & DE-RU & FI-IT & FI-RU & AVG \\
\hline 
\rowcolor{llmclirGrey} \multicolumn{11}{l}{\textit{First-stage retrieval}\rule{0pt}{8pt}} \\
\\[-8pt]

NV-Embed-v2 & 0.286 & 0.450 & 0.324 & 0.422 & 0.148 & \textbf{0.404} & 0.287 & 0.342 & \textbf{0.244} & 0.323 \\
BM25-DT & \textbf{0.413} & 0.396 & 0.255 & 0.485 & 0.301 & 0.282 & 0.216 & 0.245 & 0.179 & 0.308 \\
Hybrid & 0.407 & \textbf{0.480} & \textbf{0.336} & \textbf{0.516} & \textbf{0.372} & 0.406 & \textbf{0.303} & \textbf{0.344} & 0.221 & \textbf{0.376} \\

\cdashline{1-11}[.4pt/1pt]\noalign{\vskip 0.5ex}
\textcolor{gray}{NV-Embed-v2 \textit{(oracle)}} & 
\textcolor{gray}{0.499} & \textcolor{gray}{0.738} & \textcolor{gray}{0.545} &
\textcolor{gray}{0.683} & \textcolor{gray}{0.485} & \textcolor{gray}{0.684} &
\textcolor{gray}{0.539} & \textcolor{gray}{0.624} & \textcolor{gray}{0.477} & \textcolor{gray}{0.586} \\
\textcolor{gray}{BM25-DT \textit{(oracle)}} & 
\textcolor{gray}{0.613} & \textcolor{gray}{0.686} & \textcolor{gray}{0.571} &
\textcolor{gray}{0.716} & \textcolor{gray}{0.617} & \textcolor{gray}{0.533} &
\textcolor{gray}{0.481} & \textcolor{gray}{0.446} & \textcolor{gray}{0.378} & \textcolor{gray}{0.560} \\

\textcolor{gray}{Hybrid \textit{(oracle)}} & \textcolor{gray}{0.648} & \textcolor{gray}{0.776} & \textcolor{gray}{0.657} & \textcolor{gray}{0.758} & \textcolor{gray}{0.665} & \textcolor{gray}{0.702} & \textcolor{gray}{0.614} & \textcolor{gray}{0.625} & \textcolor{gray}{0.505} & \textcolor{gray}{0.661} \\ 
\\[-8pt]

\rowcolor{llmclirGrey} \multicolumn{11}{l}{\textit{Listwise Reranking (Retriever: hybrid)}\rule{0pt}{8pt}} \\
\\[-8pt]
RankZephyr (OG) & 0.399 & 0.479 & \textbf{0.393} & 0.509 & 0.386 & 0.398 & 0.359 & 0.320 & \textbf{0.211} & 0.384 \\ 
RankZephyr (DT) & \textbf{0.468$^*$} & \textbf{0.489} & \textbf{0.393} & \textbf{0.528} & \textbf{0.462$^*$} & \textbf{0.440} & \textbf{0.387$^*$} & \textbf{0.319} & 0.206 & \textbf{0.410} \\ 

\rowcolor{llmclirGrey} \multicolumn{11}{l}{\textit{Pairwise Reranking (Retriever: hybrid)}\rule{0pt}{8pt}} \\
\\[-8pt]
Llama-3.1-8B-Instruct (OG) & \textbf{0.447$^*$} &  0.502 & \textbf{0.407$^*$} &  0.513 &  0.406 & 0.451$^*$ &  0.367 & 0.389$^*$ & \textbf{0.301$^*$} & 0.420 \\
Llama-3.1-8B-Instruct (DT) & 0.463$^*$ & 0.508$^*$ &  0.381 & \textbf{0.535$^*$} & \textbf{0.436$^*$} & \textbf{0.471$^*$} &  0.339 & 0.388$^*$ &  0.294 & \textbf{0.424} \\
Aya-101 (OG) & 0.443$^*$ & \textbf{0.509} & 0.363 & 0.522 & 0.417$^*$ & 0.459$^*$ & \textbf{0.373$^*$} & \textbf{0.397$^*$} & 0.298 & 0.420 \\
Aya-101 (DT) & 0.436$^*$ &  0.501 &  0.374 &  0.524 & 0.421$^*$ & 0.457$^*$ & 0.369$^*$ &  0.359 &  0.247 & 0.410 \\

\bottomrule
\end{tabular}
\caption{MAP scores of hybrid retriever and listwise reranking results on CLEF 2003, with the best performance for each language pair and retrieval stage marked in \textbf{Bold}. $^*$: statistically significant difference to the first-stage retriever (paired t-test, $p<0.05$).}
\label{tab:clef_hybrid}
\end{table*}


\setlength{\tabcolsep}{12pt}
\begin{table*}[ht!]
\centering
\small 
\begin{tabular}{llllll}
\toprule
\textbf{} & EN-HA & EN-SO & EN-SW & EN-YO & AVG \\
\hline 

\rowcolor{llmclirGrey} \multicolumn{6}{l}{\textit{First-stage retrieval}\rule{0pt}{8pt}} \\
\\[-8pt]
M3 & \textbf{0.388} & 0.351 & \textbf{0.402} & 0.425 & 0.392 \\
BM25-DT & 0.214 & 0.236 & 0.233 & 0.445 & 0.282 \\
Hybrid & 0.382 & \textbf{0.358} & 0.377 & \textbf{0.497} & \textbf{0.403} \\ 

\cdashline{1-6}[.4pt/1pt]\noalign{\vskip 0.5ex}

\textcolor{gray}{M3 \textit{(oracle)}} & \textcolor{gray}{0.744} & \textcolor{gray}{0.687} & \textcolor{gray}{0.792} & \textcolor{gray}{0.793} & \textcolor{gray}{0.754}  \\
\textcolor{gray}{BM25-DT \textit{(oracle)}} & \textcolor{gray}{0.586} & \textcolor{gray}{0.561} & \textcolor{gray}{0.611} & \textcolor{gray}{0.826} & \textcolor{gray}{0.646}  \\
\textcolor{gray}{Hybrid \textit{(oracle)}} & \textcolor{gray}{0.586} & \textcolor{gray}{0.687} & \textcolor{gray}{0.792} & \textcolor{gray}{0.793} & \textcolor{gray}{0.715} \\ 

\\[-8pt]

\rowcolor{llmclirGrey} \multicolumn{6}{l}{\textit{Listwise Reranking (Retriever: hybrid)}\rule{0pt}{8pt}} \\
\\[-8pt]
RankZephyr (OG) & 0.359 & 0.336 & 0.390 & 0.477 & 0.390 \\ 
RankZephyr (DT) & \textbf{0.474$^*$} & \textbf{0.462$^*$} & \textbf{0.474$^*$} & \textbf{0.571$^*$} & \textbf{0.495} \\ 

\rowcolor{llmclirGrey} \multicolumn{6}{l}{\textit{Pairwise Reranking (Retriever: hybrid)}\rule{0pt}{8pt}} \\
\\[-8pt]
Llama-3.1-8B-Instruct (OG) & 0.397$^*$ & 0.367 & 0.416$^*$ & 0.517$^*$ & 0.424 \\
Llama-3.1-8B-Instruct (DT) & 0.417$^*$ & \textbf{0.405$^*$} & 0.418$^*$ & \textbf{0.540$^*$} & 0.445 \\
Aya-101 (OG) & 0.455$^*$ & 0.402$^*$ & \textbf{0.427$^*$} & 0.536$^*$ & \textbf{0.455} \\
Aya-101 (DT) & \textbf{0.456$^*$} & 0.404$^*$ & 0.422$^*$ & 0.535$^*$ & 0.454 \\

\bottomrule
\end{tabular}
\caption{nDCG@20 scores of hybrid retrieval and reranking on the CIRAL dataset, with the best performance for each language pair and retrieval stage marked in \textbf{Bold}. $^*$: statistically significant difference to the first-stage retriever (paired t-test, $p<0.05$).}
\label{tab:ciral_hybrid}
\end{table*}

\section{Hybrid Retrieval Experiments}
\label{sec:hybrid}
In addition to our BM25 and bi-encoder retrieval experiments, we also evaluate a hybrid retrieval approach in which LLMs rerank fused top-100 rankings using reciprocal rank fusion \citep{cormack2009reciprocal}.\footnote{The union of the top-100 document sets of BM25 and bi-encoder is larger than 100. 
For a fair comparison, we also limit the number of documents of the fused ranking to 100 documents.} 
In the following, we limit our analysis to open-source LLMs. The retrieval and reranking results are shown in Tables~\ref{tab:clef_hybrid}~and~\ref{tab:ciral_hybrid}. 

\paragraph{Oracle results.} With regard to the best possible reranking performance achievable, we find mixed results. On CLEF, we notice that hybrid retrieval improves the reranking potential (i.e., oracle score) to 0.611 MAP, while both reranking the input rankings of NV-Embedd-v2 and BM25-DT can at best yield MAP values of 0.586 and 0.560. However, on CIRAL we find that the best possible reranking results of the hybrid model falls below the performance of the M3 retriever. This may be explained by the fact that the gap between the bi-encoder and lexical retriever is much larger on CIRAL (0.392 vs. 0.282) than on CLEF (0.323 vs. 0.308). 
Interestingly, despite the large performance gap between both prerankers on CIRAL, the hybrid model still brings slight performance improvements (with a MAP score of 0.403).

\paragraph{Listwise reranking results.} On listwise reranking on CLEF, we find that the performance of RankZephyr improves from 0.342 (see Table~\ref{tab:clef_list}) to 0.384 without translation (OG), and from 0.368 to 0.410 when documents are translated (DT). Similar gains can be seen on CIRAL (see  Table~\ref{tab:ciral_list}), where the MAP scores improve from 0.327 to 0.390 with original language documents (OG), and from 0.407 to 0.495 with translated documents. These results are consistent with our results presented in Section~\ref{sec:results} and show that improvements in retrieval translate to improvements in reranking.

\paragraph{Pairwise reranking results.} Different from the listwise reranking results, we find that the  improvements resulting from document translation diminish for pairwise rerankers. Both Llama-3.1-8B-Instruct and Aya-101 perform substantially better compared to reranking only the results of NV-Embedd-v2 and M3 for CLEF and CIRAL (see Tables~\ref{tab:clef_pair}~and~\ref{tab:ciral_pair}). On CLEF, the best performance is achieved by Llama-3.1-8B (DT), whereas RankZephyr (DT) performs best on CIRAL. 

\setlength{\tabcolsep}{3.8pt}
\begin{table*}[!ht]
\small
\centering
\begin{tabular}{l r r l r}
\toprule
\textbf{Model} & \textbf{Parameters} & \textbf{\#Lang.} & \textbf{Unsupported Languages} & \textbf{Emb. Dim.} \\
\midrule  
mGTE \citep{zhang-etal-2024-mgte} & 305M & 75 & HA & 768 \\
RepLlama \citep{repllama} & 7B & 1 & FI, DE, IT, RU, HA, SW, SO, YO & 4096 \\
M3 \citep{chen-etal-2024-m3} & 560M & 173 & HA & 1024 \\
E5 \citep{wang2024multilinguale5textembeddings} & 560M & 94 & YO & 1024 \\
NV-Embedd-v2 \citep{lee2025nvembed} & 7.85B & 1 & FI, DE, IT, RU, HA, SW, SO, YO & 4096 \\

\midrule

RankZephyr \citep{pradeep2023rankzephyreffectiverobustzeroshot} & 7.24B & 1 & FI, DE, IT, RU, HA, SW, SO, YO & - \\
RankGPT$_{3.5}$ \citep{sun-etal-2023-chatgpt} & - & - & - & - \\
RankGPT$_{4.1}$ \citep{sun-etal-2023-chatgpt} & - & - & - & - \\
Aya-101 \citep{ustun-etal-2024-aya} & 12.9B & 101 & None & - \\
Llama-3.1-8B-Instruct \citep{touvron2023llama} & 8.03B & 8 & FI, RU, HA, SW, SO, YO & - \\
\bottomrule
\end{tabular}
\caption{Retriever and reranker model information based on HuggingFace model cards \citep{wolf2020huggingfacestransformersstateoftheartnatural}. For each model, we list the number of parameters, number languages it was trained on, and which of the CIRAL and CLEF languages are not supported. For bi-encoders we also report the embedding size.}
\label{tab:model_stats}
\end{table*}

\section{Model Information}
Information about the retriever and reranker models used in our study is summarized in Table \ref{tab:model_stats}.
For retrievers, RepLlama is fine-tuned from Llama-2-7B \cite{touvron2023llama} using LoRA \cite{hu2021loralowrankadaptationlarge} on the MS MARCO Passage Ranking \cite{bajaj2018msmarcohumangenerated} training split for one epoch,
while E5 is initialized from XLM-RoBERTa-large \citep{conneau-etal-2020-unsupervised} and trained on a mixture of multilingual datasets covering 100 languages.
NV-Embed-v2 model is built upon decoder-only Mistral-7B-v0.1 \cite{jiang2023mistral7b}.

For rerankers, RankZephyr is based on the Zephyr-7B-{\ss} model \cite{tunstall2023zephyrdirectdistillationlm}. We use the RankZephyr Full version, which is distilled from RankGPT$_{3.5}$ (100K training queries) and RankGPT$_{4}$ (5K queries). The training corpus is limited to monolingual English data.

\end{document}